\documentclass[11pt]{article}

\usepackage[margin=1in]{geometry}
\usepackage{amsmath}
\usepackage{amssymb}
\usepackage{graphicx}
\usepackage{booktabs}
\usepackage{adjustbox}
\usepackage{tikz}
\usetikzlibrary{positioning,arrows.meta,calc,fit,backgrounds}
\usepackage[numbers,sort&compress]{natbib}
\usepackage{hyperref}
\usepackage{authblk}
\usepackage{caption}
\usepackage{xcolor}

\hypersetup{
    colorlinks=true,
    linkcolor=blue,
    citecolor=blue,
    urlcolor=blue
}

\title{\textbf{Memisis: Orchestrating and Evaluating Synthetic Data for Tabular Health Datasets}}

\author[1]{Nitish Nagesh}
\author[1]{Pengbao Zhou}
\author[2]{Atchuth Naveen Chilaparasetti}
\author[1]{Tu Nguyen}
\author[1]{Yajat Nagaraj Kiran}
\author[1]{Arshia Harish Puthran}
\author[1]{Muhjaazee Love}
\author[1]{Aadi Sharma}
\author[1]{Mahdi Bagheri}
\author[1]{Ian Harris}
\author[1]{Amir M. Rahmani}

\affil[1]{University of California, Irvine, CA, USA}
\affil[2]{PathAI, USA}

\date{}

\begin{document}

\maketitle

\begin{abstract}
Synthetic data is widely used in healthcare to create datasets that preserve statistical properties of real data without exposing sensitive patient information. Generating and evaluating synthetic data across privacy, utility, and fairness dimensions is crucial for enabling high-quality data availability in downstream prediction tasks and clinical decision making. We present \textbf{Memisis}, a tool that orchestrates and evaluates synthetic data by leveraging existing synthesis libraries, large language models (LLMs), and state-of-the-art evaluation metrics. Our tool creates a unified workflow for data generation, validation, and evaluation. Users can control training size, training epochs, and the number of synthetic rows to sample. Beyond manual configuration, an interactive agent mode allows users to specify data generation goals in natural language, and the tool orchestrates the full pipeline by invoking existing synthesizers while performing the requisite evaluation. For the demo, we use an open-source schizophrenia dataset with protected attributes related to race and gender, evaluate six synthesizers spanning GANs, VAEs, diffusion models, and normalizing flows, and use a local LLM to orchestrate the workflow. The system affords users flexibility and control over the data generation and evaluation process.
\end{abstract}

\section{Introduction}

Real-world data in healthcare has sensitive information and has associated privacy concerns~\citep{giuffre2023harnessing,mostly2020whybias}. Synthetic data offers a way to perform data augmentation and substitution that thereby improves access to high-quality data for downstream training and evaluation. Existing synthetic data generation platforms span a range of approaches, including generic generators~\citep{ibm2024watsonsynthetic}, LLM-based applications~\citep{argilla2024syntheticdatagenerator,nvidia2024llamasynthetic,nagesh2025faircausesyn,nagesh2025fairtabgen}, cohort-specific simulators~\citep{walonoski2018synthea}, and AI partners for data access and augmentation~\citep{callender2025synthcraft}. Existing libraries such as Synthcity~\citep{qian2023synthcity} and Synthetic Data Vault (SDV)~\citep{sdv-dev} demand machine learning expertise, commercial platforms such as MOSTLY AI~\citep{mostly2021syntheticaiml}, Gretel Navigator~\citep{gretel2025navigator}, and Tonic Fabricate~\citep{tonic2025fabricate} offer limited transparency, and LLM-based frameworks like SynthCraft~\citep{callender2025synthcraft} reduce the interaction barrier but still evaluate fairness downstream rather than enforcing it at generation time. No existing system allows a user to negotiate fairness, utility, and privacy as jointly optimized objectives during synthesis, with a verifiable audit trail. Racial and intersectional bias in clinical algorithms leads to systematic misdiagnosis~\citep{obermeyer2019dissecting,gara2012influence,gara2019naturalistic,olbert2018meta}, and fairness must be assessed at individual-group and intersectional levels~\citep{buolamwini2018gender,mehrabi2021survey,bird2020fairlearn}. Agentic workflows enable these trade-offs to be negotiated during synthesis rather than resolved post hoc~\citep{gorenshtein2025agents}. We present \textbf{Memisis}, an interactive multi-agent system that exposes schema detection, synthesizer selection, sampling, and multi-metric evaluation through natural language, and that treats fairness, utility, and privacy as jointly negotiated constraints at generation time.

\section{Methodology}

\begin{figure}[t]
    \centering
    \includegraphics[width=0.85\linewidth]{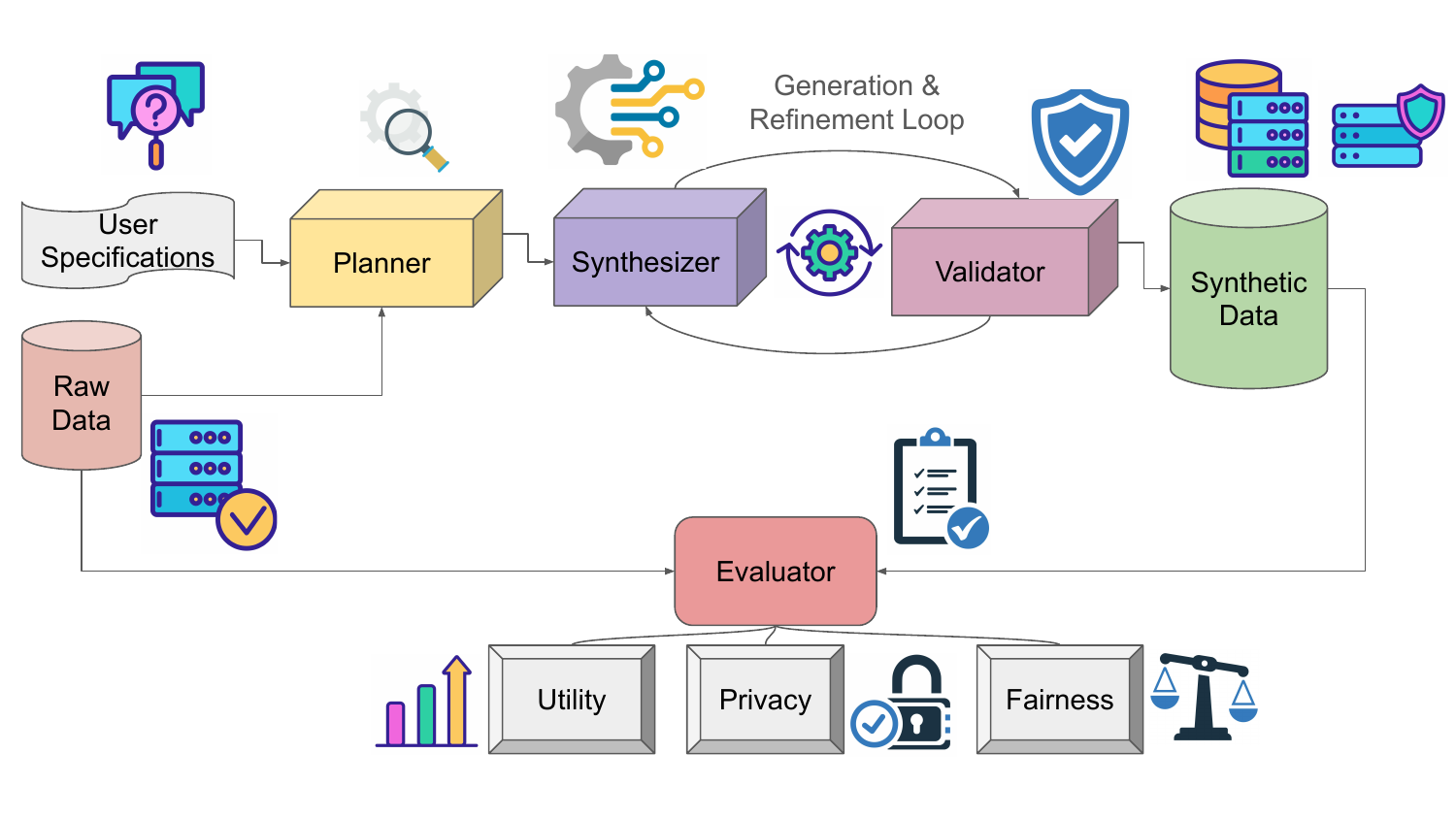}
    \caption{Overall workflow: user specifications and raw data flow through the Planner into a Synthesizer--Validator generation-and-refinement loop. The resulting synthetic data is jointly evaluated with the real data across utility, fairness, and privacy dimensions.}
    \label{fig:overall}
\end{figure}

Memisis integrates planning, synthesis, quality control, and evaluation into a unified pipeline. As shown in Figure~\ref{fig:overall}, user specifications and raw data are ingested by the Planner, which orchestrates a Synthesizer and a Validator in an iterative refinement loop. The Evaluator then assesses three dimensions: utility, fairness, and privacy. Fidelity metrics include average Jensen--Shannon Distance (JSD) and Wasserstein Distance (WD); fairness metrics cover disparate treatment, disparate impact, demographic parity ratio, and equalized odds ratio.

\begin{figure}[t]
    \centering
    \includegraphics[width=0.85\linewidth]{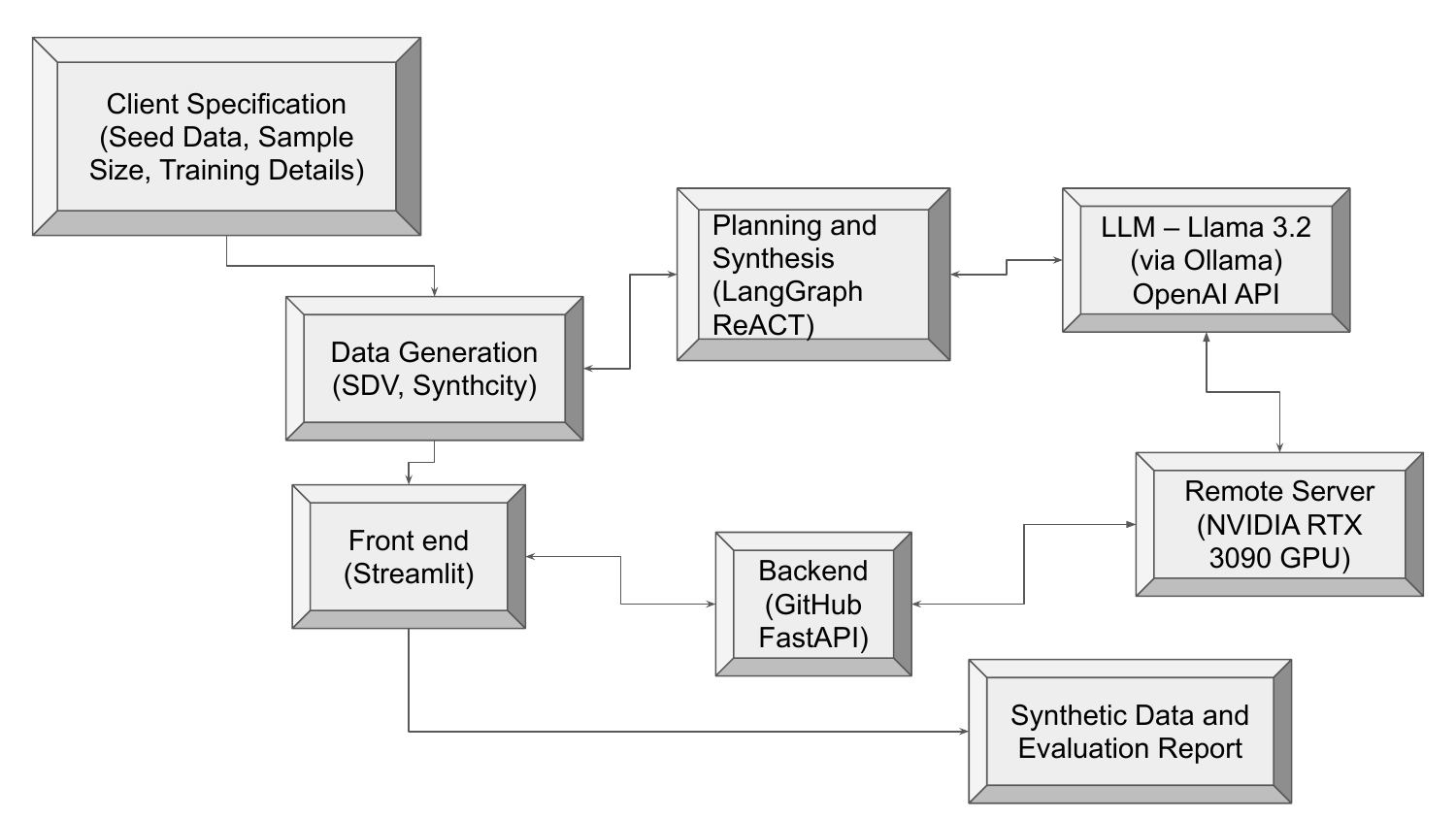}
    \caption{Tool architecture: client specifications drive LangGraph ReACT-based planning and synthesis via SDV and Synthcity generators. A Streamlit front end communicates with a FastAPI back end; LLM inference is served by Llama~3.2 via Ollama on an NVIDIA RTX~3090 GPU.}
    \label{fig:tool_stack}
\end{figure}

The tool architecture is presented in Figure~\ref{fig:tool_stack}. Synthesizers from the Synthetic Data Vault (SDV)~\citep{sdv-dev} and Synthcity~\citep{qian2023synthcity} span GANs, VAEs, diffusion models, and normalizing flows, and are trained on a configurable subset of real data. Agent orchestration uses a LangGraph ReACT framework with six specialist agents --- Schema, Generator, Fairness, Utility, Privacy, and Provenance --- where the Fairness Agent holds veto authority over the Generator Agent and can trigger regeneration when subgroup thresholds fail. The Orchestrator plans an execution graph in natural language while deterministic code carries it out; tools are exposed via the Model Context Protocol~\citep{anthropic2024mcp}. The TSTR (Train Synthetic, Test Real) paradigm validates distributional fidelity, followed by group-level fairness computation. In the generic mode, users set explicit parameters; in the agent mode, Memisis accepts natural language goals and orchestrates the full pipeline autonomously. The front end is built with Streamlit (deployed on Streamlit Cloud via GitHub) and the back end with FastAPI, with LLM inference via Llama~3.2 through Ollama on an NVIDIA RTX~3090 GPU.

\section{Results}

We evaluate on an open-source schizophrenia tabular dataset from OpenML~\citep{bischl2025openml} documenting intersectional disparities in mental health diagnosis~\citep{gara2012influence,gara2019naturalistic,olbert2018meta}. The dataset includes \texttt{Race}, \texttt{Sex}, clinical fields, and a binary \texttt{Diagnosis} label (schizophrenia vs.\ affective disorder). We train on 1{,}000 real rows, sample 500 synthetic rows, and run each synthesizer for 20 epochs.

\begin{table}[h]
    \centering\small
    \begin{tabular}{lrrrr}
    \toprule
     & Train Acc. & Test Acc. & Specificity & Sensitivity \\
    \midrule
    Original & 0.93 & 0.94 & 0.93 & 0.94 \\
    CTGAN    & 0.84 & 0.81 & 0.76 & 0.85 \\
    TVAE     & 0.85 & 0.84 & 0.83 & 0.85 \\
    DDPM     & 0.93 & 0.94 & 0.93 & 0.94 \\
    PATEGAN  & 0.57 & 0.58 & 0.68 & 0.49 \\
    RTVAE    & 0.84 & 0.80 & 0.73 & 0.85 \\
    NFLOW    & 0.77 & 0.73 & 0.85 & 0.55 \\
    \bottomrule
    \end{tabular}
    \caption{Utility metrics (TSTR paradigm). DDPM matches the original data across all metrics.}
    \label{tab:performance}
\end{table}

\begin{table*}[t]
    \centering\footnotesize
    \begin{tabular}{lrrrrrrr}
    \toprule
    & Original & CTGAN & TVAE & DDPM & PATEGAN & RTVAE & NFLOW \\
    \midrule
    \multicolumn{8}{l}{\textit{Relative FPR by Race (normalized to White)}} \\
    \midrule
    Asian    & 0.77 & 1.2  & 0.39 & 0.37 & 0.55 & 1.3  & 1.3  \\
    Black    & 2.9  & 1.8  & 1.3  & 2.5  & 0.48 & 1.8  & 1.7  \\
    Hispanic & 1.3  & 1.3  & 2.5  & 1.9  & 0.68 & 0.73 & 0.96 \\
    White    & 1.0  & 1.0  & 1.0  & 1.0  & 1.0  & 1.0  & 1.0  \\
    \midrule
    \multicolumn{8}{l}{\textit{Relative intersectional FPR by Race $\times$ Sex (normalized to White.Male)}} \\
    \midrule
    Asian.F    & 0.00 & 0.88 & 0.00 & 0.36 & 1.1  & 1.1  & 1.2  \\
    Asian.M    & 1.4  & 1.8  & 0.67 & 0.00 & 0.00 & 0.64 & 1.1  \\
    Black.F    & 1.4  & 1.8  & 0.88 & 0.68 & 0.66 & 1.4  & 1.1  \\
    Black.M    & 5.2  & 2.0  & 1.8  & 3.0  & 0.65 & 1.2  & 1.8  \\
    Hispanic.F & 2.0  & 0.92 & 1.7  & 0.76 & 0.51 & 0.29 & 0.89 \\
    Hispanic.M & 0.00 & 2.1  & 2.4  & 0.85 & 1.4  & 0.95 & 0.76 \\
    White.F    & 0.69 & 1.1  & 0.73 & 0.07 & 1.6  & 0.59 & 0.73 \\
    White.M    & 1.0  & 1.0  & 1.0  & 1.0  & 1.0  & 1.0  & 1.0  \\
    \bottomrule
    \end{tabular}
    \caption{Fairness evaluation across synthesizers. Values above 1.0 indicate higher misdiagnosis rates relative to the reference group (White for racial FPR; White.Male for intersectional FPR). F\,=\,Female, M\,=\,Male.}
    \label{tab:fairness}
\end{table*}

We benchmark six synthesizers spanning GANs, VAEs, diffusion models, and normalizing flows on a schizophrenia misdiagnosis dataset, evaluating (i)~utility and fidelity --- accuracy, sensitivity, and specificity; and (ii)~fairness --- relative false positive rate (FPR) by racial group and by intersectional race-and-gender group. DDPM achieves the highest test sensitivity and specificity, most closely matching the original data across all synthesizers --- GANs (CTGAN, PATEGAN), VAEs (TVAE, RTVAE), and NFLOW. DDPM also exhibits the smallest deviation in racial FPR from the original profile, indicating high distributional fidelity, and preserves the intersectional bias in the original dataset (Black.Male relative FPR of 3.0 vs.\ 5.2 in original). PATEGAN achieves the lowest absolute FPR values across racial groups, but at the cost of substantially reduced utility. Across all synthesizers, the false positive rate --- the rate of misdiagnosis --- remains consistently higher for Black men than for White men, mirroring the original dataset and corroborated by prior literature~\citep{gara2012influence,gara2019naturalistic,olbert2018meta}. The small representation of Asian and Hispanic subgroups warrants deeper investigation into fairness estimate reliability for these groups.

\section{Discussion}

Evaluating bias before deploying ML models for clinical diagnosis is essential~\citep{obermeyer2019dissecting,bird2020fairlearn}, particularly for underrepresented groups where real data is scarce and predictive models fail hardest~\citep{mehrabi2021survey}. The tool is currently used by a small research community; near-term goals include expanding the metric suite and supported datasets, with longer-term plans for HL7/HIPAA compliance and user studies with clinical data scientists. A key insight: the novelty in this space rarely lies in a new generator --- it lies in the coordination layer and control structures built around existing ones.

\end{document}